\documentclass[letterpaper]{article} 
\usepackage{aaai24}  
\usepackage{times}  
\usepackage{helvet}  
\usepackage{courier}  
\usepackage[hyphens]{url}  
\usepackage{graphicx} 
\usepackage{natbib}  
\usepackage{caption} 
\usepackage{multirow}
\usepackage{bbding}
\usepackage{amssymb}
\usepackage[mathscr]{eucal}
\usepackage[linesnumbered,ruled,lined,boxed,commentsnumbered]{algorithm2e}
\usepackage{algpseudocode}

\newcommand{\ie}{\textit{i.e.}}
\newcommand{\eg}{\textit{e.g.}}
\usepackage{siunitx}
\usepackage{amsmath}
\usepackage{newfloat}
\usepackage{listings}
\DeclareCaptionStyle{ruled}{labelfont=normalfont,labelsep=colon,strut=off} 
\title{Negative Pre-aware for Noisy Cross-modal Matching}
\author{
    Xu Zhang\textsuperscript{\rm 1}\footnotemark[2],
    Hao Li\textsuperscript{\rm 1}\footnotemark[2],
    Mang Ye\textsuperscript{\rm 2}\footnotemark[1]
}

\affiliations{
    \textsuperscript{\rm 1}School of Computer Science and Engineering, University of Electronic Science and Technology of China\\
    \textsuperscript{\rm 2}School of Computer Science, Wuhan University\\
    \{xuzhang.xoe, 18th.leolee, mangye16\}@gmail.com
}

\usepackage{bibentry}

\begin{document}
\maketitle
\footnotetext[1]{Corresponding author.}
\footnotetext[2]{These authors contributed equally to this work.}
\begin{abstract}
Cross-modal noise-robust learning is a challenging task since noisy correspondence is hard to recognize and rectify. Due to the cumulative and unavoidable negative impact of unresolved noise, existing methods cannot maintain a stable performance when the noise increases. In this paper, we present a novel Negative Pre-aware Cross-modal (NPC) matching solution for large visual-language model fine-tuning on noisy downstream tasks. It is featured in two aspects: (1) For noise recognition and resistance, previous methods usually directly filter out a noise subset, we propose to estimate the negative impact of each sample. It does not need additional correction mechanisms that may predict unreliable correction results, leading to self-reinforcing error. We assign a confidence weight to each sample according to its negative impact in the training process. This adaptively adjusts the contribution of each sample to avoid noisy accumulation. (2) For maintaining stable performance with increasing noise, we utilize the memorization effect of DNNs by maintaining a memory bank. Specifically, we apply GMM to select high-confident clean samples as the memory entry, where the memory entry is used to estimate the negative impact of each sample. Since clean samples are easier distinguished by GMM with increasing noise, the memory bank can still maintain high quality at a high noise ratio. Compared to the correction mechanism focusing on noise samples, memory bank-based estimation is more robust, which makes the model performance stable on noisy datasets. Extensive experiments demonstrate that our method significantly improves matching accuracy and performance stability at increasing noise ratio. Our approach also surpasses the state-of-the-art methods by a large margin. The code is available at: \url{https://github.com/ZhangXu0963/NPC}.
\end{abstract}

\section{Introduction}
Cross-modal matching aims to align different modalities (\eg, text and image) within a common space and pair them based on similarity score. With the explosion of multimedia data, cross-modal matching has gained traction in both industry and academia, \eg, text-to-image generation~\cite{c:tigan, c:CogView}, image captioning~\cite{r:V2t, r:s2t, c:eff}, and visual question answering~\cite{r:Revive, c:symbolic}. 
\begin{figure}[h]
\centering
\includegraphics[width=\columnwidth, trim=0 0 1000 340, clip]{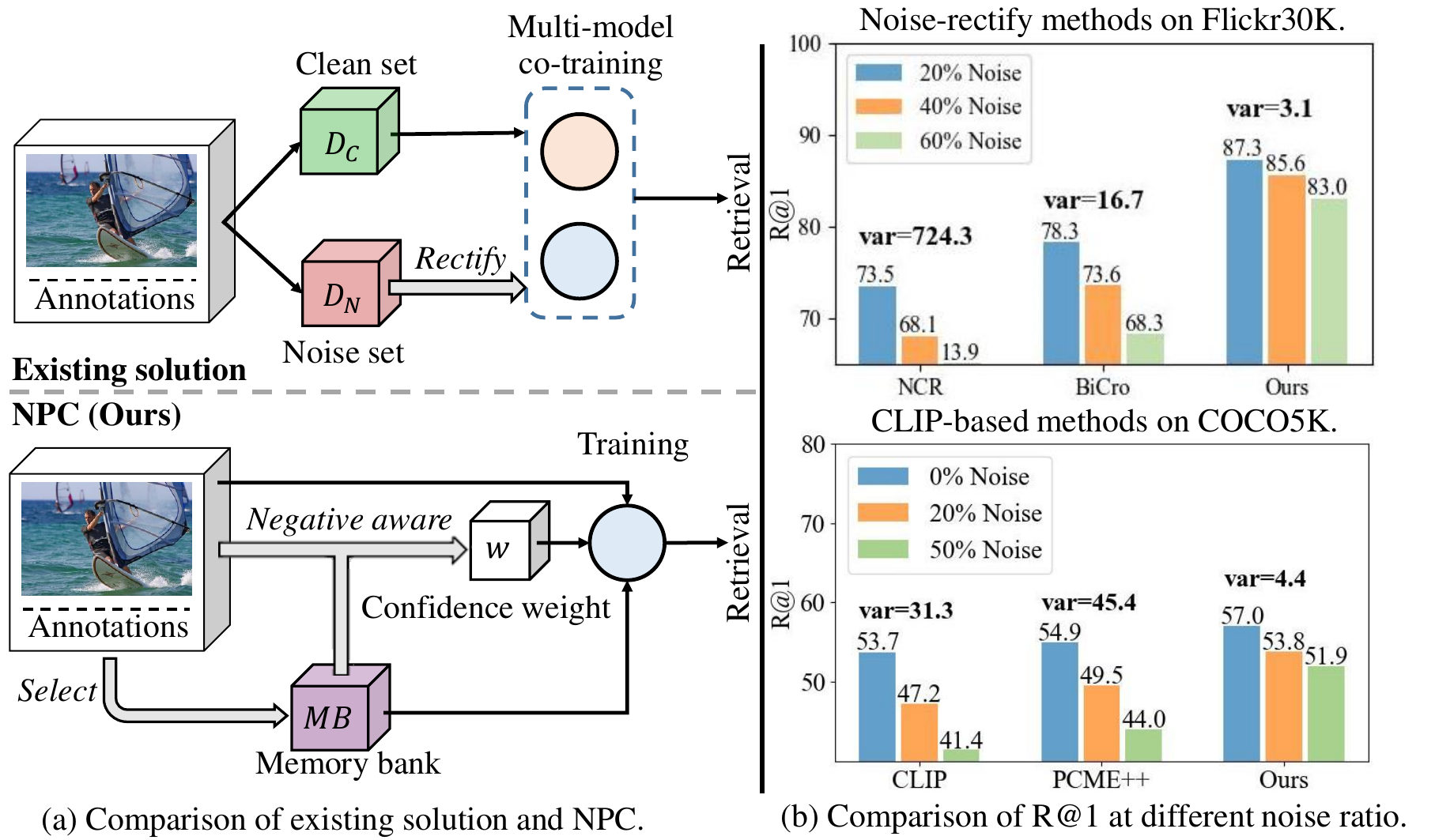} 
\caption{(a) Existing solution vs NPC. (b) Sensitivity of noise-robust learning methods and CLIP-based methods with increasing noise.  
}
\label{f1}
\end{figure}

These works have achieved promising performance by training on large-scale datasets. However, it is expensive to obtain a well-annotated dataset in practical scenarios. The manual-annotated datasets, \eg, MSCOCO~\cite{c:coco}, Flickr30K~\cite{r:f30k}, and  Conceptual Captions~\cite{c:CC}, incorporate a significant number of inaccurate descriptions, namely noisy correspondence. Unlike noisy label in classification tasks, the noise here is mismatched cross-modal pairs which is more difficult to deal with, since involves both visual and textual modeling. Therefore, a series of approaches~\cite{r:NCR,c:Bicro,c:MSCN} following the noise-rectify paradigm have been developed to counter the negative impact of the noise. These methods typically filter out the noise subset from the original training set, and address the noise issue through label correction. Nevertheless, the inherent flaw of the noise-rectify paradigm cannot maintain the performance stability in the existence of severe noise. As shown in Fig.~\ref{f1}(b), we compare the performance of different methods using R@1 metric, including noise-rectify based approaches~\cite{r:NCR, c:Bicro}, the CLIP-based approaches~\cite{c:clip, pcme++} and our approach. We employ variance ($var$) of R@1 at different noise ratio to illustrate the ``performance stability''. Obviously, noise-rectify based methods exhibit unstable performance with a considerably larger variance than ours. Additionally, CLIP-based methods also lack consistent performance with increasing noise, even though CLIP is a powerful pre-trained model. Most existing noise-rectify paradigms rely on collaborative rectifying with multiple models.
Since the limitation of the rectifying mechanism, the matching performance under high-noise is unstable.
In these works~\cite{r:NCR,c:Bicro}, the new labels are entirely estimated by DNN models. With high noise ratio, some indistinguishable noise correspondences are prone to be directly learned and remembered by DNNs, ultimately leading to a dramatic drop in performance under high-noise.
Existing methods emphasize the ``discrimination learning'' ability but ignore the ``stability''. 
In our opinion, two essential abilities are required for noise-robust learning for large visual-language model fine-tuning on noisy downstream tasks: 1) \textit{the ability to distinguish between noisy and clean samples}, 2) \textit{maintain the stability of ``discriminative learning'' with increasing noise}. 

To address aforementioned challenges, we propose a novel approach named Negative Pre-aware Cross-modal (NPC) matching. NPC adopts a unique Negative Pre-aware (NP) paradigm for robust learning. Unlike previous paradigms that mainly focus on noise filtering or correction, the NP paradigm adaptively assesses the potential negative impact of each sample before the model learning (see Fig.~\ref{f1}(a)). 
DNNs tends to prioritize learning easy samples over noisy and challenging ones~\cite{c:closerlook, c:robustel}. With gradually fitting noise samples, the model begins to generate incorrect predictions~\cite{c:Earlylearning}. 
In other words, once the model learned a noise pair, fitting certain specific clean samples becomes more challenging. These clean samples usually have images or texts that are similar to the noise pair. 

Inspired by this phenomenon, our NPC uses easy-distinguishable clean samples to estimate negative impacts. 
We rigorously choose a reliable clean subset from the training data by using Gaussian Mixture Model~\cite{r:Dividemix,r:study} to fit the loss distribution of each pair. 
And high-confident clean samples are maintained in a Memory Bank ($MB$), which is used to assist the model in estimating negative impact before to fully model training. A small confidence weight will be assigned to high-negative samples. 

The main contributions are summarized as follows:
\begin{itemize}
    \item We highlight the challenge with large visual-language model fine-tuning on noisy downstream tasks, \ie, how to achieve robust learning in cross-modal matching with the increasing amount of noise.  
    \item We introduce the Negative Pre-aware Cross-modal (NPC) matching paradigm by establishing a memory bank for negative impact estimation. We employ the assistance of memory entries to allocate confidence weights ($w$) to the samples. These components constitute the cornerstones to achieving stable and highly noise-resistant performance.

    \item Extensive experiments are conducted on two manual-annotated datasets and a real-world dataset, showcasing the NPC's superiority over the state-of-the-art methods. Moreover, with the increasing noise, both quantitative and qualitative results affirm that NPC demonstrates notably higher performance stability compared to previous methods.
\end{itemize}
\section{Related Works}
\subsection{Image-text Matching}
Typical image-text matching methods align data from different modalities to measure similarity. Early works~\cite{r:vse++,c:pie,r:wangliwei,c:dual} mainly focus on global alignments. 
Some prior works~\cite{c:scan, c:vsrn,c:SGRAF,c:vsl} adopt attention mechanisms to achieve fine-grained local alignments.
Subsequently, many works~\cite{c:PCME,pcme++,r:DAA} devote to modeling the many-to-many relationships in image-caption pairs. 

Recently, with the success of transformer-based vision and language models~\cite{c:vit, c:bert}, vision-language pre-training (VLP) models, such as CLIP~\cite{c:clip}, have shown strong performance in multiple cross-modal tasks~\cite{c:IRRA,c:CoBIT}. Although VLP possesses impressive zero-shot ability, it still reveals vulnerabilities in training with noisy datasets on specific downstream tasks. In this paper, we employ CLIP as our backbone and introduce an anti-noise learning strategy.
\subsection{Cross-modal Noise-robust Learning}
Huang et al.~\cite{r:NCR} first tackle noise correspondences, which consider mismatched cross-modal pairs instead of incorrect annotations. Since then, several approaches~\cite{c:MSCN, c:Bicro, r:tip21, rc:tifs20} have developed the noise-rectify process in various cross-modal tasks. They can be categorized into two groups: noise correction and noise re-weighting. Noise correction methods achieve robust learning by correcting the similarity~\cite{c:MSCN} or correspondence label~\cite{r:NCR,c:Bicro} of noise pairs. The noise re-weighting methods~\cite{c:DECL} degrade the contribution of noise samples to achieve robust learning. 
All these methods require splitting a noise subset from the original training dataset. Subsequently, they proceed with rectification within this subset. Nonetheless, as noise increases, the imprecise subset division and inaccurate rectification can amplify adverse effects. Different from these works, we sidestep the problem by forecasting per-sample negative impact following the novel NP paradigm.
\begin{figure*}[t]
\centering
\includegraphics[width=0.9\textwidth, trim=0 0 200 240, clip]{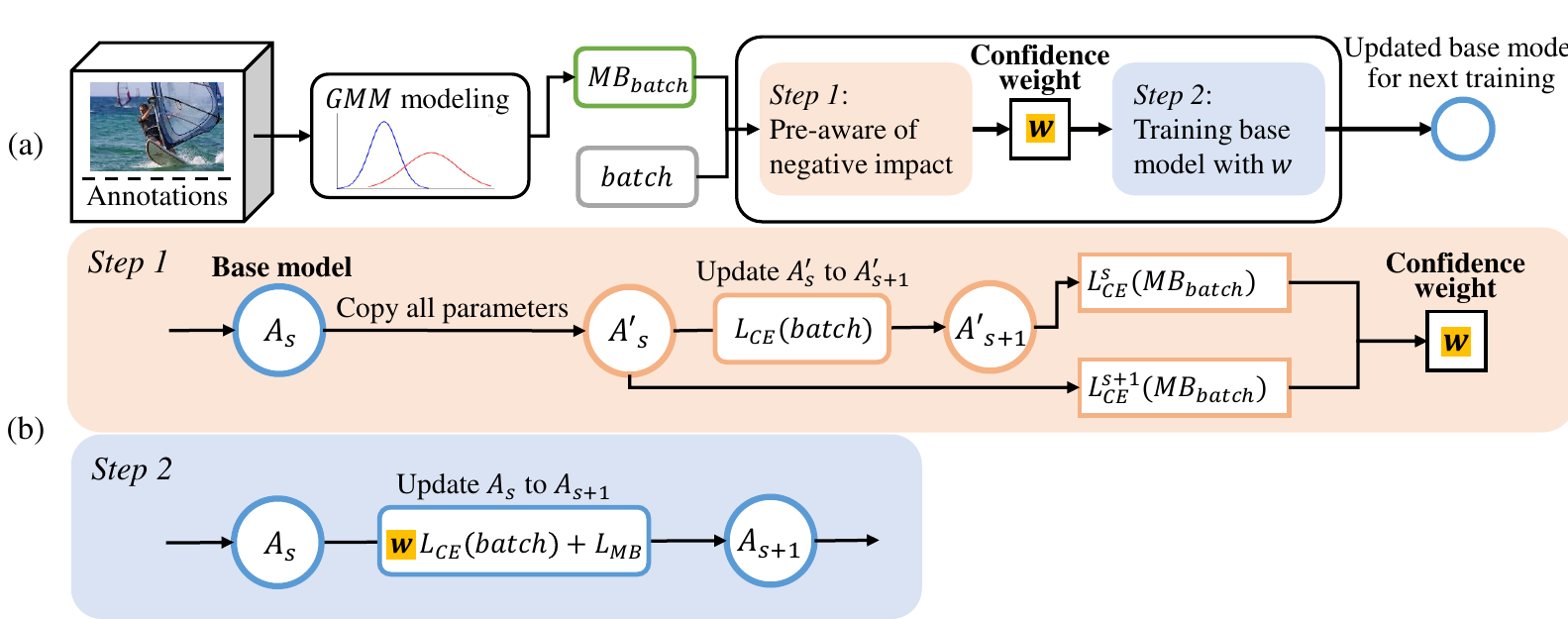} 
\caption{
(a) \textbf{Illustrating the NPC training pipeline.} Given a batch of image-text pairs, we select their corresponding memory entries from a strict clean set divided by GMM as inputs. Then we optimize the base model in two steps: the first step aims to estimate the negative impact and obtain per-sample confidence weight $w$. The second step is training the base model with $w$. (b) \textbf{Illustrating two training steps.} We first share all parameters of the base model $A_s$ to its siamese model $A'_s$. Then we train the model $A'_s$ on the batch samples, obtaining the model $A'_{s+1}$. The negative impact of each sample can be calculated by comparing its loss of corresponding memory entry on $A'_s$ and $A'_{s+1}$. If the loss on $A'_{s+1}$ is higher than it on $A'_s$, this means the sample brings a negative impact to the model, and we will give it a low confidence weight. After the negative-aware process, the model $A_s$ will be trained with the re-weight samples and memory bank, generating the robust target model $A_{s+1}$.}
\label{f2}
\end{figure*}

\section{Proposed Method}
\subsection{Preliminary}
\subsubsection{Problem Definition.} Given a dataset $D=\{(I_i, T_i)\}_{i=1}^{N}$, where $(I_i, T_i)$ is the $i^{th}$ image-text pair, and $N$ denotes the data size. The goal of image-text matching is to align the visual and textual modalities within a shared space to calculate the similarity following Eq.~\ref{e1},
\begin{equation}\label{e1}
    S(I_i, T_j)=\frac{f(I_i)\cdot g(T_j)}{\Vert f(I_i)\Vert\cdot\Vert g(T_j)\Vert},
\end{equation}
where $f(\cdot)$ and $g(\cdot)$ serve as feature extractors for two modalities. Generally, positive pairs exhibit higher similarity scores, whereas negative pairs show lower similarity scores. 

\subsubsection{Revisiting CLIP-based Solution.}
With the emergence of the VLP model CLIP~\cite{c:clip} as a compelling option for cross-modal downstream tasks, we employ CLIP as the pre-trained backbone for the proposed NPC approach. CLIP enhances visual and textual feature extractors through the minimization of the symmetric cross-entropy loss $\mathcal{L}_{CE}(I_i, T_i)$, which is defined as follows:
\begin{equation} \label{e2}
    \begin{aligned}
        \mathcal{L}_{CE}(I_i, T_i&)=CE(I_i, T_i)+CE(T_i, I_i), \\
        CE(x_i, y_i)&=-\frac{1}{N}\sum_{i=1}^N \log\bigg(\frac{\exp(S(x_i, y_i))}{\sum_{j=1}^N \exp(S(x_i, y_j))}\bigg).\\
    \end{aligned}
\end{equation}
However, Eq.~\ref{e2} works effectively based on the assumption that $(I_i,T_i)$ constitutes a positive pair. Yet, when $(I_i,T_i)$ is a noise correspondence, relying solely on Eq.~\ref{e2} can lead to a substantial detrimental impact on the model. 

Fig.~\ref{f1} provides a clear visual representation, demonstrating that when the noise ratio rises from 20\% to 60\%, the CLIP's R@1 performance experiences a steep decline from 82.3\% to 66.3\%. 
Therefore, the NPC approach is introduced to enhance the stability and robustness of pre-trained models in tackling noise challenges.
The training pipeline, depicted in Fig.~\ref{f2}, comprises the two main components that will be elaborated upon in the subsequent section. 


\subsection{Memory Bank Construction}
We propose to estimate the negative impact of each sample brought to the model during the training process. A direct approach is to evaluate the performance changes of the model before and after training. Limited by the high cost of evaluating on the test set, we construct corresponding evaluation entries for each sample, which together form a Memory Bank ($MB$). Concretely, we select these entries from a reliable clean set to guarantee the accuracy of evaluation.  
\begin{figure}[t]
\centering
\includegraphics[width=\columnwidth, trim=0 0 650 360, clip]{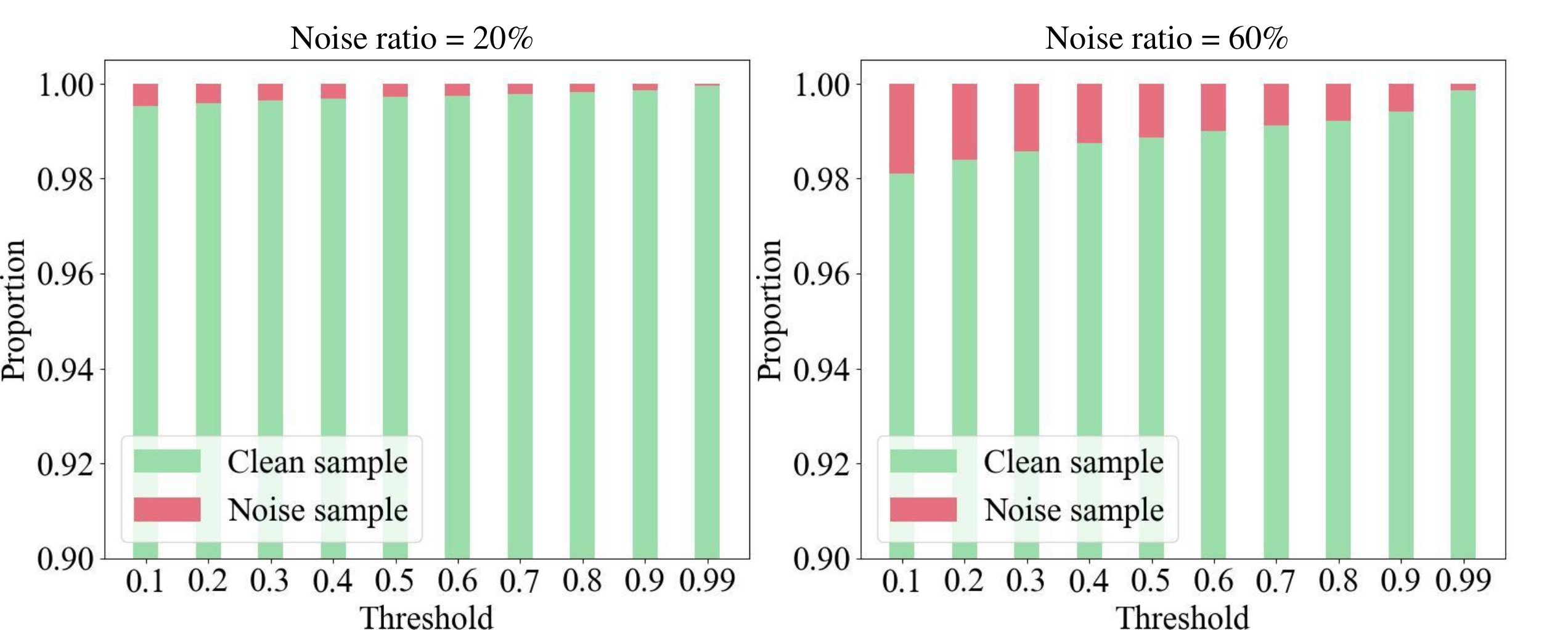} 
\caption{The proportion of noise and clean samples in the clean set, obtained through GMM at different thresholds $\tau$. 
Generally, samples with the posterior probability of $p_i \ge \tau$ are included in the clean set. Inevitably, there are some noise samples in it. The threshold $\tau=0.99$ ensures that the clean set selected from either the low (e.g. 20\%) or the high (e.g. 60\%) noise ratio training set is virtually noise-free.}
\label{f3}
\end{figure}
Since DNN tends to learn the easy patterns before noisy and hard patterns~\cite{c:closerlook, c:robustel}, clean samples typically exhibit lower loss values than the noisy or hard ones. Based on this, we leverage the difference in loss distribution among samples to discern clean pairs. Following NCR~\cite{r:NCR}, we utilize a two-component Gaussian Mixture Model to fit the distribution of per-sample loss in the training dataset:
\begin{equation}\label{e3}
p(z|\theta)=\sum_{k=1}^K\alpha_k\phi(z|\theta_k),
\end{equation}
where $\alpha_k$ represents the mixture coefficient, and $\phi(z|\theta_k)$ denotes the probability density of the $k^{th}$ component. The posterior probability computed by Eq.~\ref{e4} serves as the clean probability $p_i$ for the $i^{th}$ sample.
\begin{equation}\label{e4}
p_i=p(\theta_k|z_i)=\frac{p(\theta_k)p(z_i|\theta_k)}{p(z_i)}.
\end{equation}
Here, $\theta_k$ refers to the Gaussian component with a lower mean.
The samples with $p_i\ge\tau$ are considered clean, as indicated in Eq.~\ref{e5}.
\begin{equation}\label{e5}
    D_c = \{(I_j,T_j)\vert p_j\ge \tau\}.
\end{equation}
Fig.~\ref{f3} illustrates the proportion of noise and clean samples in the selected dataset $D_c$ with varied threshold $\tau$. We perform the strict selection using $\tau=0.99$ to obtain the clean set $D_c$, practically devoid of noise. Strict selection is a prerequisite to ensure the reliability of the memory bank.


Then, we need to select evaluation entries in the strict clean set for each sample to construct the memory bank. For each pair $(I_i,T_i)$ in the training set, we first select an image-text pair $(I_i^I,T_i^I)$ from $D_c$ for $I_i$, where the image in this pair ($I_i^I$) exhibits the highest cosine similarity (Eq.~\ref{e2}) with $I_i$. Similarly, we also choose an image-text pair  $(I_i^T,T_i^T)$ for $T_i$. The constructed memory bank can be defined as $MB=\{(I_i^I,T_i^I), (I_i^T,T_i^T)\}_{i=1}^N$. 

\subsection{Pre-aware of the Negative Impact}
An intuitive fact is that when the model learns a noisy sample, its prediction accuracy of related clean samples will be declined. Therefore, after a sample is trained, we can determine its negative impact degree through the model performance on related clean samples. To estimate the negative impact of each sample, we have built the related clean evaluation entries for each sample, which together form a Memory Bank ($MB$).

During the batch with the size of $m$ training shown in Fig.~\ref{f2}, both batch data and their corresponding memory entry set $MB_{batch}=\{b_1,b_2,\dots,b_m\}$ are inputted into the model simultaneously. In the initial phase of each batch training, the base model $A$ shares all parameters with $A'$. It's worth noting that the models $A$ and $A'$ update separately and independently. The purpose of $A'$ is to perceive the negative impact of each sample in the batch by assessing the performance changes of the model on $MB_{batch}$ after training. We utilize the loss to denote the performance, \ie, the low loss almost means the model performs well on $MB_{batch}$. For the image-text pair $(I_k,T_k)$, the losses of its evaluation entry $b_k$ on both i2t and t2i can be computed by:
\begin{equation}\label{e6}
    \begin{aligned}        p_k&=CE(I_k^I,T_k^I)+CE(I_k^T,T_k^T),\\     
        q_k&=CE(T_k^I, I_k^I)+CE(T_k^T, I_k^T).\\
    \end{aligned}
\end{equation}
Denote the model before and after training as $A'_s$ and $A'_{s+1}$, respectively. The performance change of the model after the sample $(I_k, T_k)$ trained can be calculated by:
\begin{equation}\label{e7}
    r_k=\frac{1}{2}\bigg(\frac{p_{k}^s}{p_{k}^{s+1}}+\frac{q_k^s}{q_{k}^{s+1}}\bigg).
\end{equation}
When $r_k<1$, \ie, the loss $p_k$ and $q_k$ increase after training. It means that the model's ability on predicting the correspondence of the clean pair related to the sample $(I_k, T_k)$ is declined after training it. Thus, $(I_k, T_k)$ has a negative impact on the model $A'$. We utilize the confidence weight $w_k$ to quantify the negative impact of the pair $(I_k,T_k)$ following Eq.~\ref{e8}. The sample with high negative impact (\ie, low $r_k$) should correspond to a small confidence weight $w_k$.
\begin{equation}\label{e8}
    w_k=
    \begin{cases}
        \hfil \tanh(r_k)&, r_k<1\\
        \hfil 1&, otherwise\\
    \end{cases}
\end{equation}
The sample with $r_k<1$ will bring a negative impact to the model. Thus, we will assign the confidence weight $w_k<1$ computed by a tangent function for it. Similarly, for the samples with $r_k \ge 1$, we will assign the confidence weight $w_k=1$. So far, in the batch, we can estimate the negative impact of each sample on the base model $A$.

\begin{table*}[t]\small
\caption{Image-Text Matching on MSCOCO 1K and Flickr30K.}
\label{t1}
\resizebox{0.97\textwidth}{62mm}{
\begin{tabular}{c|c|cccccc|cccccc}
\hline
\multirow{3}{*}{noise} & \multirow{3}{*}{method} & \multicolumn{6}{c|}{MSCOCO 1K}                    & \multicolumn{6}{c}{Flickr30K}                     \\ \cline{3-14} 
                       &                         & \multicolumn{3}{c}{image-to-text} & \multicolumn{3}{c|}{text-to-image} & \multicolumn{3}{c}{image-to-text} & \multicolumn{3}{c}{text-to-image} \\
                       &                         & R@1     & R@5     & R@10   & R@1     & R@5     & R@10    & R@1     & R@5    & R@10    & R@1     & R@5    & R@10    \\ \hline
\multirow{7}{*}{0\%}  & SCAN         & 69.2          & 93.6          & 97.6          & 56.0          & 86.5          & 93.5          & 67.4          & 90.3          & 95.8          & 48.6          & 77.7          & 85.2          \\
                      & SAF          & 76.1          & 95.4          & 98.3          & 61.8          & 89.4          & 95.3          & 73.7          & 93.3          & 96.3          & 56.1          & 81.5          & 88.0          \\
                      & NCR          & 78.7          & 95.8          & 98.5          & 63.3          & 90.4          & 95.8          & 77.3          & 94.0          & 97.5          & 59.6          & 84.4          & 89.9          \\
                      & DECL         & 79.1          & 96.3          & 98.7          & 63.3          & 90.1          & 95.6          & 79.8          & 94.9          & 97.4          & 59.5          & 83.9          & 89.5          \\
                      & BiCro       & 79.1          & 96.4          & 98.6          & 63.8          & 90.4          & 96.0          & 81.7          & 95.3          & 98.4          & 61.6          & 85.6          & 90.8          \\
                      & CLIP         & 79.9          & 95.1          & 98.1          & 65.0          & 90.3          & 98.1          & 86.2          & 97.6          & 99.2          & 72.9          & 92.3          & 96.0          \\
                      & \textbf{NPC} & \textbf{82.2} & \textbf{96.5} & \textbf{98.7} & \textbf{68.3} & \textbf{92.0} & \textbf{98.7} & \textbf{87.9} & \textbf{98.1} & \textbf{99.4} & \textbf{75.0} & \textbf{93.7} & \textbf{97.2} \\ \hline
\multirow{7}{*}{20\%} & SCAN         & 62.2          & 90.0          & 96.1          & 46.2          & 80.8          & 89.2          & 58.5          & 81.0          & 90.8          & 35.5          & 65.0          & 75.2          \\
                      & SAF          & 71.5          & 94.0          & 97.5          & 57.8          & 86.4          & 91.9          & 62.8          & 88.7          & 93.9          & 49.7          & 73.6          & 78.0          \\
                      & NCR          & 77.7          & 95.5          & 98.2          & 62.5          & 89.3          & 95.3          & 73.5          & 93.2          & 96.6          & 56.9          & 82.4          & 88.5          \\
                      & DECL         & 77.5          & 95.9          & 98.4          & 61.7          & 89.3          & 95.4          & 77.5          & 93.8          & 97.0          & 56.1          & 81.8          & 88.5          \\
                      & BiCro       & 78.8          & \textbf{96.1} & \textbf{98.6} & 63.7          & 90.3          & 95.7          & 78.1          & 94.4          & 97.5          & 60.4          & 84.4          & 89.9          \\
                      & CLIP         & 75.0          & 93.1          & 97.2          & 58.7          & 86.1          & 97.2          & 82.3          & 95.5          & 98.3          & 66.0          & 88.5          & 93.5          \\
                      & \textbf{NPC} & \textbf{79.9} & 95.9          & 98.4          & \textbf{66.3} & \textbf{90.8} & \textbf{98.4} & \textbf{87.3} & \textbf{97.5} & \textbf{98.8} & \textbf{72.9} & \textbf{92.1} & \textbf{95.8} \\ \hline
\multirow{7}{*}{40\%} & SCAN         & 42.9          & 74.6          & 85.1          & 24.2          & 52.6          & 63.8          & 26.0          & 57.4          & 71.8          & 17.8          & 40.5          & 51.4          \\
                      & SAF          & 13.5          & 43.8          & 48.2          & 16.0          & 39.0          & 50.8          & 7.4           & 19.6          & 26.7          & 4.4           & 12.2          & 17.0          \\
                      & NCR          & 74.7          & 94.6          & 98.0          & 59.6          & 88.1          & 94.7          & 68.1          & 89.6          & 94.8          & 51.4          & 78.4          & 84.8          \\
                      & DECL         & 75.6          & 95.5          & \textbf{98.3} & 59.5          & 88.3          & 94.8          & 72.7          & 92.3          & 95.4          & 53.4          & 79.4          & 86.4          \\
                      & BiCro       & 77.0          & \textbf{95.9} & \textbf{98.3} & 61.8          & 89.2          & 94.9          & 74.6          & 92.7          & 96.2          & 55.5          & 81.1          & 87.4          \\
                      & CLIP         & 70.7          & 91.7          & 96.2          & 54.7          & 83.4          & 96.2          & 76.2          & 93.3          & 96.5          & 59.4          & 85.0          & 90.9          \\
                      & \textbf{NPC} & \textbf{79.4} & 95.1          & \textbf{98.3} & \textbf{65.0} & \textbf{90.1} & \textbf{98.3} & \textbf{85.6} & \textbf{97.5} & \textbf{98.4} & \textbf{71.3} & \textbf{91.3} & \textbf{95.3} \\ \hline
\multirow{7}{*}{60\%} & SCAN         & 29.9          & 60.9          & 74.8          & 0.9           & 2.4           & 4.1           & 13.6          & 36.5          & 50.3          & 4.8           & 13.6          & 19.8          \\
                      & SAF          & 0.1           & 0.5           & 0.7           & 0.8           & 3.5           & 6.3           & 0.1           & 1.5           & 2.8           & 0.4           & 1.2           & 2.3           \\
                      & NCR          & 0.1           & 0.3           & 0.4           & 0.1           & 0.5           & 1.0           & 13.9          & 37.7          & 50.5          & 11.0          & 30.1          & 41.4          \\
                      & DECL         & 73.0          & 94.2          & \textbf{97.9} & 57.0          & 86.6          & 93.8          & 65.2          & 88.4          & 94.0          & 46.8          & 74.0          & 82.2          \\
                      & BiCro       & 73.9          & \textbf{94.4} & 97.8          & 58.3          & 87.2          & 93.9          & 67.6          & 90.8          & 94.4          & 51.2          & 77.6          & 84.7          \\
                      & CLIP         & 67.0          & 88.8          & 95.0          & 49.7          & 79.6          & 95.0          & 66.3          & 87.3          & 93.0            & 52.1          & 78.8          & 87.4          \\
                      & \textbf{NPC} & \textbf{78.2} & \textbf{94.4} & 97.7          & \textbf{63.1} & \textbf{89.0} & \textbf{97.7} & \textbf{83.0} & \textbf{95.9} & \textbf{98.6} & \textbf{68.1} & \textbf{89.6} & \textbf{94.2} \\ \hline
\end{tabular}}
\end{table*}

\subsection{Re-training}
After negative impact evaluation, we need to re-train the model $A$ to get the robust target model $A_{s+1}$.
To avoid the detriment of the samples with a negative impact on the base model $A$, we re-weight the symmetric cross-entropy loss:
\begin{equation} \label{e9}
    \begin{aligned}
    \mathcal L_{RCE}=\frac{1}{m}\sum_{k=1}^m w_k \mathcal{L}_{CE}(I_k, T_k).
    \end{aligned}
\end{equation}
For these detrimental samples, the labels are not reliable. To further mitigate the detriment of these unreliable labels to the model, we employ the related memory entries to help the model learn the correct correspondences (Eq.~\ref{e10}).
\begin{equation} \label{e10}
    \mathcal L_{MB}=\frac{1}{m}\sum_{k=1}^m \left[ \mathcal{L}_{CE}(I_k^I, T_k^I) + \mathcal{L}_{CE}(I_k^T, T_k^T)\right].
\end{equation}
Thus, the total objective function in the re-training process can be denoted as:
\begin{equation} \label{e11}
    \mathcal L_{total}=\mathcal L_{RCE}+\mathcal L_{MB}.
\end{equation}
\section{Experiments}
\subsection{Experimental Setting}
\subsubsection{Datasets and Evaluation Metrics.} The proposed NPC is evaluated on three benchmark datasets, MSCOCO~\cite{c:coco}, Flickr30K~\cite{r:f30k}, and CC120K:
\begin{itemize}
    \item MSCOCO contains 123,287 images with 5 annotated captions per image. Following previous works~\cite{r:NCR}, we use 113,287 images for training, 5,000 images for validation, and 5,000 images for testing.
    \item Flickr30K contains 31,783 images with 5 annotated texts per image. Following previous works~\cite{r:NCR}, we use 29,783 images for training, 1,000 images for validation, and 1,000 images for testing.
    \item CC120K. We randomly sample a subset from the real-world dataset Conceptual Captions~\cite{c:CC}. This dataset is harvested from the Internet, with about 3\%-20\% incorrect image-text pairs. CC120K contains 120, 851 with a single caption per image. In our experiment, we use 118,851 images for training, 1,000 images for validation, and 1,000 images for testing.
\end{itemize}
The widely-used metric Recall@K (R@K) is used to evaluate the performance of image-text matching with K=1, 5, and 10. The variance ($var$) of R@1 at different noise ratios is used to evaluate the approaches' performance stability, with lower $var$ indicating higher stability.

\subsubsection{Implementation Details.} 
NPC can enhance noise resistance and stability in various cross-modal matching models.
In this paper, the CLIP~\cite{c:clip} with ViT-B/32 is implemented as a baseline. Both baseline and NPC are trained on a single RTX 3090 GPU optimized by AdamW~\cite{c:adamw}. We start training CLIP and NPC with learning rates $5e-7$ and $2e-7$ with a weight decay of 0.2. In all experiments, we train the model for 5 epochs with a mini-batch size of 256, and the hyperparameter $\tau$ is set to 0.99.
\begin{figure}[h]
\centering
\includegraphics[width=\columnwidth, trim=12 5 50 40, clip]{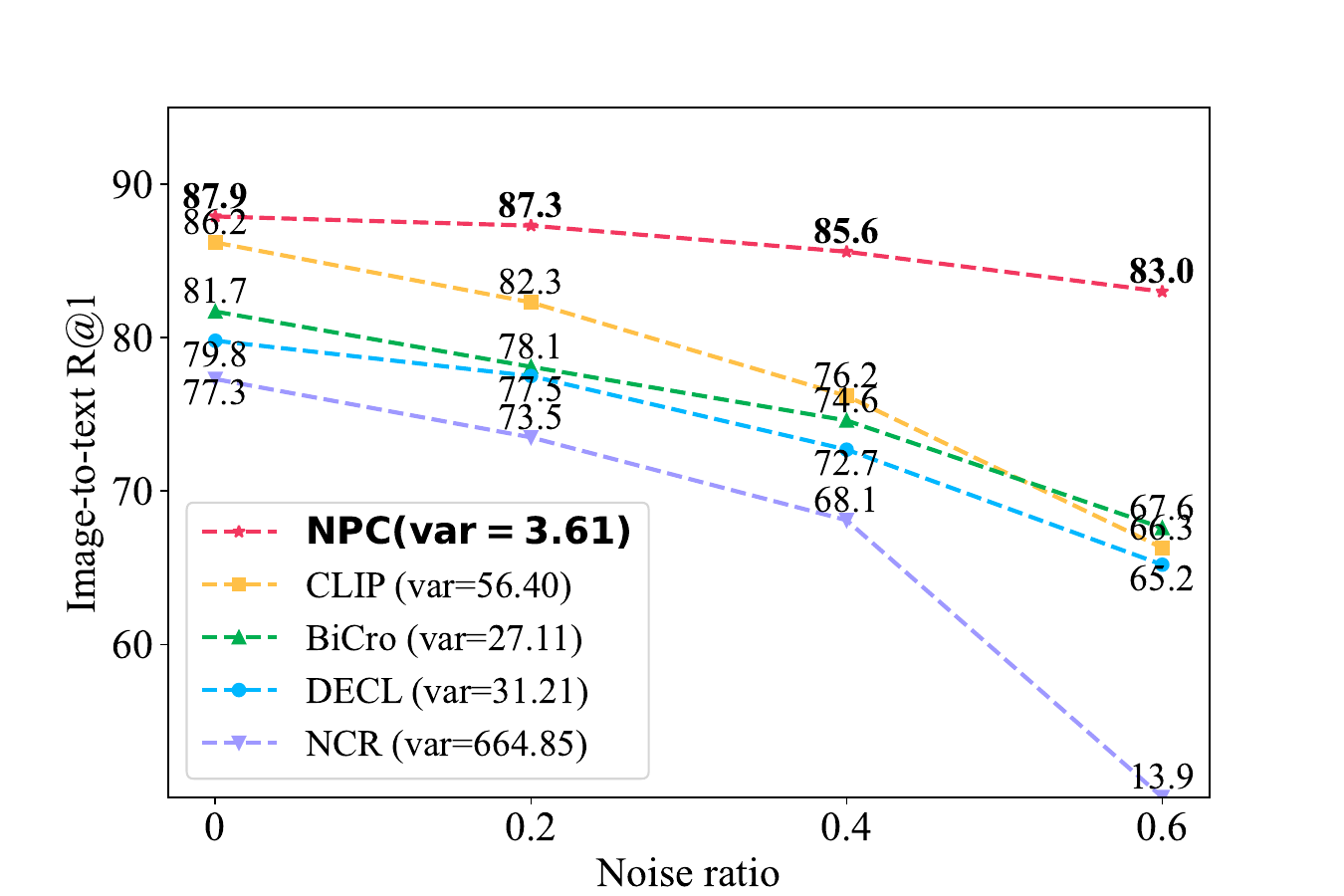} 
\caption{Variation and variance ($var$) of matching performance at different noise ratio.}
\label{f4}
\end{figure}
\subsection{Comparison with State of the Arts}
\subsubsection{Quantitative Comparison.} 
To illustrate the effectiveness, we compare NPC with various approaches, including general cross-modal matching methods SCAN~\cite{c:scan}, SAF~\cite{c:SGRAF}, noise-robust learning methods NCR~\cite{r:NCR}, DECL~\cite{c:DECL},  BiCro~\cite{c:Bicro}, and CLIP with fine-tuning~\cite{c:clip}. It is worth noting that CLIP is the baseline of our method. The results are shown in Table~\ref{t1}.


 It shows that NPC significantly outperforms all methods across all noise ratios. Notably, on Flickr30K with 60\% noise ratio, NPC outperforms the current state-of-the-art approach BiCro with a large R@1 performance gap. To be specific, the R@1 performance of NPC is 15.4\% higher than BiCro on image-text matching (i2t), as well as 16.9\% higher than BiCro on text-to-image matching (t2i). Compared to the baseline CLIP, NPC has achieved immense improvement in all metrics and benchmarks. Furthermore, as the noise ratio increases, the performance gap between NPC and baseline becomes larger. For instance, on the MSCOCO 1K set, when the noise ratio ranges from 0\% to 60\%, the R@1 performance gap between NPC and baseline separately increases from 2.3\% to 11.2\% on i2t, and 3.3\% to 13.4\% on t2i. This phenomenon is powerful to prove the effectiveness of NPC on robust learning.

\subsubsection{Stability Comparison.}
To further explore the superiority of NPC on stable learning, we illustrate the R@1 change curves of different methods under different noise ratios in Fig.~\ref{f4}. We can observe that NPC outperforms all other methods in all noise ratios. Meanwhile, as the noise ratio increases, the performance decline of NPC is significantly smaller than that of other methods.
Furthermore, we calculate the variance of each method on different noise ratios to quantify the stability of the methods. NPC shows remarkable stability with only 3.61\% variance, outperforming all other methods with a huge gap. Compared to the baseline CLIP, NPC yields a large drop on $var$ of 52.79\%. The large decrease in variance indicates the performance stability is significantly improved by NPC.

\begin{figure*}[t]
\centering
\includegraphics[width=0.9\textwidth, trim=0 0 350 192, clip]{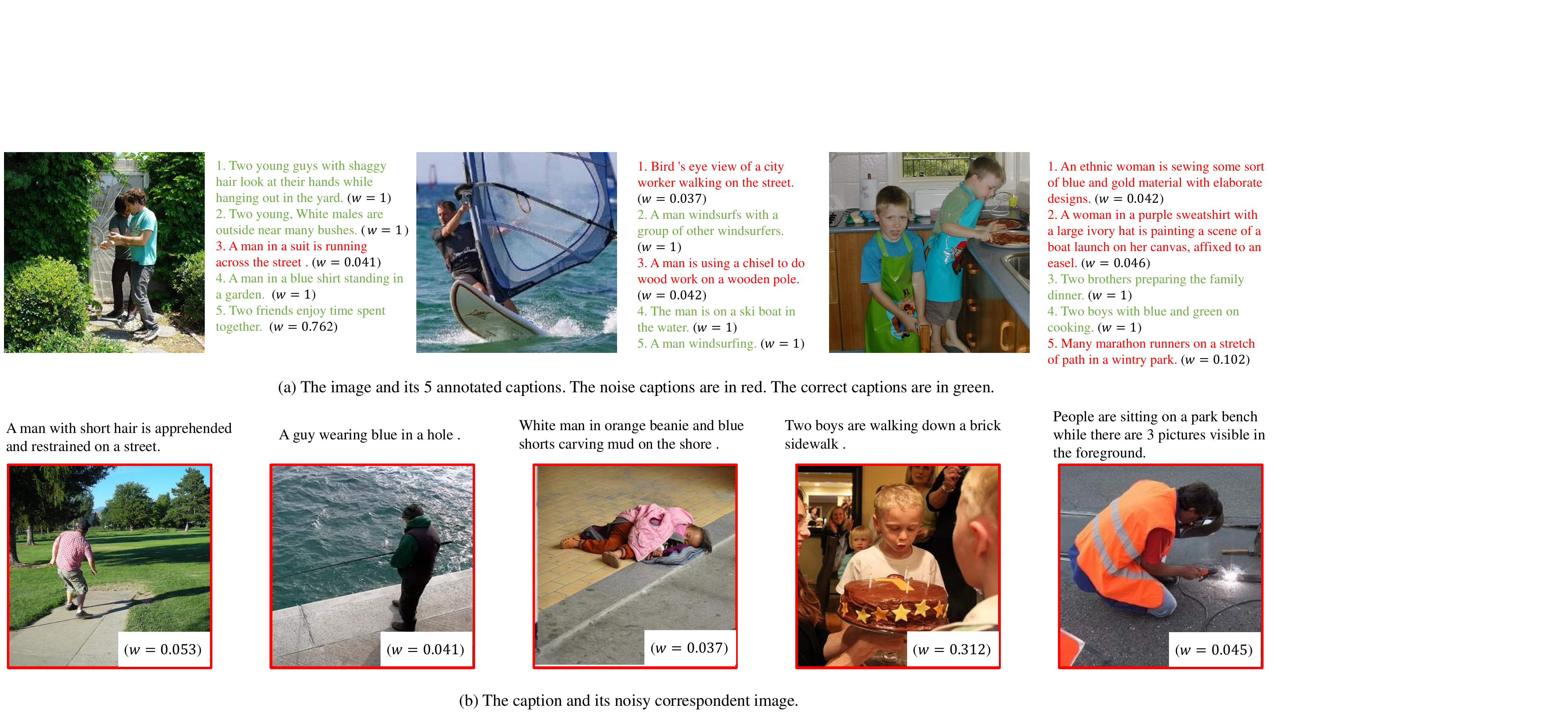} 
\caption{Some noisy correspondences in Flickr30K training set under 40\% noise. The average confidence weight ($w$) in all epochs is shown for examples. The $w$ of correctly matched pairs are obviously larger than noisy pairs.}
\label{f5}
\end{figure*}
\subsection{Comparison with ViT-B/32 Backbone Methods}
\begin{table}[t]
\caption{Comparison with baseline on CC120K.}
\label{t2}
\resizebox{\columnwidth}{8mm}{%
\begin{tabular}{c|cccccc}
\hline
\multirow{2}{*}{method} & \multicolumn{3}{c}{image-to-text}             & \multicolumn{3}{c}{text-to-image}             \\
                        & R@1           & R@5           & R@10          & R@1           & R@5           & R@10          \\ \hline
CLIP                    & 68.8          & 87.0          & 92.9          & 67.8          & 86.4          & 90.9          \\
\textbf{NPC}                     & \textbf{71.1} & \textbf{92.0} & \textbf{96.2} & \textbf{73.0} & \textbf{90.5} & \textbf{94.8} \\ \hline
\end{tabular}%
}
\end{table}
\begin{table}[t]\small
\centering
\caption{Comparison of methods with ViT-B/32 backbone on noise-free MSCOCO 5K.}
\label{t3}
\resizebox{\columnwidth}{16mm}{%
\begin{tabular}{c|cccccc}
\hline
\multirow{2}{*}{method} & \multicolumn{3}{c}{image-to-text}             & \multicolumn{3}{c}{text-to-image}             \\
                        & R@1           & R@5           & R@10          & R@1           & R@5           & R@10          \\ \hline
VSE$\infty$                     & 60.2          & 85.4          & 92.2          & 46.9          & 75.5          & 84.8          \\
PCME                 & 59.9          & 85.8          & 92.3          & 46.1          & 75.0          & 84.6          \\
PCME++                  & 61.8          & 87.0          & 93.0          & 47.9          & 76.5          & 85.4          \\
PAU                 & 63.6           & 85.2         & 92.2          & 46.8          & 74.4          & 83.7              \\
CLIP                    & 62.2          & 84.6           & 90.9         & 45.1          & 72.3      & 81.8              \\
\textbf{NPC}            & \textbf{65.4} & \textbf{87.3} & \textbf{93.1} & \textbf{48.5} & \textbf{75.4} & \textbf{84.4} \\ \hline
\end{tabular}%
}
\end{table}
In Table~\ref{t2}, we compare the NPC with baseline on the CC120K which is with real noisy correspondences. From the results, our proposed method outperforms the baseline by a considerable margin in terms of all metrics. Specifically, NPC is 2.3\% and 5.2\% higher than CLIP on i2t and t2i R@1, respectively.

For a fair comparison, we also compare the NPC to the methods with the same CLIP ViT-B/32-based backbone, including VSE$\infty$~\cite{c:VSE}, PCME~\cite{c:PCME}, PCME++~\cite{pcme++}, and PAU~\cite{r:pau}. The results on noise-free MSCOCO 5K are shown in Table~\ref{t3}. It demonstrates that NPC consistently outperforms other methods in all metrics. Besides, we also report the average R@1 of image-to-text and text-to-image of MSCOCO 1K and 5K in Table~\ref{t4} at different noise ratios. Meanwhile, the sum of R@1, R@5, and R@10 on both i2t and t2i on MSCOCO 1K is also reported. As the noise ratio increases, NPC outperforms others by larger margins, surpassing the second best model PAU by 2.0\% at 20\% noise ratio, while 2.3\% at 50\% noise ratio for 5K R@1. All these experiments effectively demonstrate the effectiveness and superiority of NPC.
\begin{table}[t]\small
\centering
\caption{Comparison of methods with ViT-B/32 backbone on noisy MSCOCO.}
\label{t4}
\resizebox{0.95\columnwidth}{24mm}{%
\begin{tabular}{c|c|ccc}
\hline
noise                 & method       & 1K R@1        & 5K R@1         & 1K RSUM        \\ \hline
\multirow{6}{*}{20\%} & VSE$\infty$          & 72.0            & 51.4           & 520.2          \\
                      & PCME      & 69.9          & 48.1           & 519.3          \\
                      & PCME++       & 70.8          & 49.5           & 522.4          \\
                      & PAU       & 71.4             & 51.7           & 521.5              \\
                      & CLIP         & 66.8          & 47.2           & 507.2           \\
                      & \textbf{NPC} & \textbf{73.1} & \textbf{53.8}  & \textbf{529.8} \\ \hline
\multirow{6}{*}{50\%} & VSE          & 38.5          & 18.4           & 390.5          \\
                      & PCME      & 65.8            & 43.0           & 505.7          \\
                      & PCME++       & 65.7          & 44.0             & 503.9          \\
                      & PAU       & 69.3             & 49.6           & 513.4              \\
                      & CLIP         & 60.9          & 41.4             & 486            \\
                      & \textbf{NPC} & \textbf{71.3} & \textbf{51.9} & \textbf{523.4} \\ \hline
\end{tabular}%
}
\end{table}

\subsection{Ablation Study}

\subsubsection{Analysis on $w$ and $\mathcal L_{MB}$.}
According to Eq.~\ref{e11}, there are two important components of confidence weight $w$ and memory bank loss $\mathcal L_{MB}$ in the re-training process. To explore the effect of each component, we exhaustively ablate them in Flickr30K with three noise ratios. The results are shown in Table~\ref{t5}.

We observe that both $w$ and $\mathcal L_{MB}$ obtain significant performance improvements in different noise ratios. They bring almost the same improvements for NPC compared with the baseline. Specifically, training with 60\% noise, the ablative NPC exceeds the baseline by 11.8\% and 6.95\% on average R@1 of image-to-text and text-to-image, indicating that $w$ and $\mathcal L_{MB}$ have independent effect of anti-noise. Moreover, the full version NPC outperforms by a much larger margin than the baseline, indicating that both components can complement each other and collaborate to achieve robust learning. The reason why $w$ and $\mathcal L_{MB}$ can achieve robust learning is that the confidence weight $w$ mitigates the degree of negative impact from the noisy sample to the model, and the memory bank loss $\mathcal L_{MB}$ can provide correct correspondences for these noisy samples.
\begin{table}[t]
\centering
\caption{Ablation study of threshold $\tau$ on Flickr30k.}
\label{t5}
\resizebox{0.95\columnwidth}{13mm}{%
\begin{tabular}{c|c|cccccc}
\hline
\multirow{2}{*}{noise} & \multirow{2}{*}{threshold $\tau$} & \multicolumn{3}{c}{image-to-text}             & \multicolumn{3}{c}{text-to-image}             \\
                       &                            & R@1           & R@5           & R@10          & R@1           & R@5           & R@10          \\ \hline
\multirow{3}{*}{0\%}   & 0.5                        & 87.2          & \textbf{98.1} & 99.2          & 74.5          & \textbf{93.7} & 96.9          \\
                       & 0.7                        & 87.6          & 97.9          & \textbf{99.4}          & 74.9          & 93.5          & 97.1          \\
                       & 0.99                       & \textbf{87.9} & \textbf{98.1} & \textbf{99.4} & \textbf{75.0} & \textbf{93.7} & \textbf{97.2} \\ \hline
\multirow{3}{*}{60\%}  & 0.5                        & 78.3          & 94.2          & 96.7          & 59.2          & 82.6          & 88.8          \\
                       & 0.7                        & 82.2          & \textbf{95.9} & 98.3          & 67.8          & 89.4          & \textbf{94.2} \\
                       & 0.99                       & \textbf{83.1} & \textbf{95.9} & 98.6          & \textbf{68.1} & \textbf{89.6} & \textbf{94.2} \\ \hline
\end{tabular}%
}
\end{table}
\begin{table}[t]
\caption{Ablation studies for $w$ and $\mathcal L_{MB}$ on Flickr30K.}
\label{t6}
\resizebox{0.45\textwidth}{22mm}{
\begin{tabular}{c|cc|cccccc}
\hline
\multirow{2}{*}{noise} & \multicolumn{2}{c|}{method} & \multicolumn{3}{c}{image-to-text} & \multicolumn{3}{c}{text-to-image} \\
                       & $w$          & $\mathcal L_{MB}$          & R@1        & R@5        & R@10       & R@1        & R@5        & R@10       \\ \hline
\multirow{4}{*}{20\%}  & \checkmark          & \checkmark              & \textbf{87.3}      & \textbf{97.5}      & \textbf{98.8}      & \textbf{72.9}      & \textbf{92.1}      & \textbf{95.8}      \\
                       & \checkmark          &                & 85.3      & 97.3      & \textbf{98.8}      & 71.8      & 91.3      & 95.2      \\
                       &            & \checkmark              & 85.4      & 97.2      & 98.6      & 71.9      & 91.4      & 95.2      \\
                       &            &                & 82.3      & 95.5      & 98.3      & 66.0      & 88.5      & 93.5      \\ \hline
\multirow{4}{*}{40\%}  & \checkmark          & \checkmark              & \textbf{85.6}      & \textbf{97.5}      & \textbf{98.4}      & \textbf{71.3}      & \textbf{91.3}      & \textbf{95.3}      \\
                       & \checkmark          &                & 79.9      & 95.5      & 97.7      & 62.4      & 85.5      & 91.1      \\
                       &            & \checkmark              & 79.0      & 95.0      & 97.5      & 62.3      & 85.2      & 91.1      \\
                       &            &                & 76.2      & 93.3      & 96.5      & 59.4      & 85.0      & 90.9      \\ \hline
\multirow{4}{*}{60\%}  & \checkmark          & \checkmark              & \textbf{83.0}      & \textbf{95.9}      & \textbf{98.6}      & \textbf{68.1}      & \textbf{89.6}      & \textbf{94.2}      \\
                       & \checkmark          &                & 78.2      & 93.5      & 96.8      & 59.0        & 82.5      & 88.4      \\
                       &            & \checkmark              & 78.0      & 93.9      & 96.6      & 59.1      & 82.3      & 88.7      \\
                       &            &                & 66.3      & 87.3      & 93.0      & 52.1      & 78.8      & 87.4      \\ \hline
\end{tabular}}
\end{table}
\subsubsection{Analysis on hyperparameter $\tau$.} $\tau$ is a very important parameter, which can control the clean degree of clean set $D_c$ in Eq.~\ref{e5} and memory bank $MB$. A smaller value of $\tau$ leads to a larger scale of $D_c$, potentially containing more noise pairs. The purity of $D_c$ directly impacts the quality of $MB$, which in turn influences the model's matching performance. To explore the impact of the selection threshold $\tau$ on the model, we report the matching performance with different $\tau$ on Flickr30K with 0\% and 60\% noise ratios in Table~\ref{t6}, respectively. 
The results show that when training with 0\% noise, the impact of varying $\tau$ on performance reduction is not noticeable. However, in the case of training with 60\% noise, performance drops by 4.8\% and 7.0\% on R@1 when $\tau$ changes from 0.99 to 0.5. It implies that a rigorous selection of $D_c$ is necessary to establish a trustworthy $MB$.

\subsection{Visualization}
To illustrate the effectiveness of NPC, we showcase examples from Flickr30K in Fig.~\ref{f5}. The average confidence weight ($w$) for each pair across five epochs is depicted. Noisy pairs consistently exhibit notably low $w$ values. Especially in Fig.~\ref{f5} (a), there is a very obvious contrast between the $w$ of the same image with correct annotations and noisy annotations. That is to say, with the support of $MB$, NPC effectively differentiates between clean and noisy correspondences. It also avoids model learning errors by assigning a small $w$ to the noisy correspondence.
\section{Conclusion}
This paper studies a novel challenge of maintaining stable performance for the noise-robust learning model as noise increases. To tackle this, a novel approach NPC is proposed. We introduce a novel NP paradigm to estimate per-sample negative impact before it is learned by the model. 
To obtain the negative impact, the memory bank of the training set is constructed by strict selection. 
To mitigate negative impact on the model, each sample is assigned a confidence weight based on the memory bank. 
Extensive experiments indicate the effectiveness of each component in our method. The NPC achieves notable enhancement in matching accuracy and performance stability compared to the state-of-the-art approach on both noise and noise-free datasets.

\noindent \textbf{Acknowledgement.} This work is partially supported by National Natural Science Foundation of China under Grants (62176188) and the Key Research and Development Program of Hubei Province (2021BAD175)

\bibliography{aaai24}
\end{document}